\documentclass{article}
\usepackage{spconf,amsmath,graphicx,amssymb}
\usepackage{booktabs}
\usepackage{subfigure}
\usepackage{diagbox}
\usepackage{array}
\usepackage{makecell}
\usepackage{float}
\usepackage{graphicx}
\usepackage{indentfirst}
\usepackage{enumitem}
\setlist{nosep, leftmargin=14pt}

\usepackage{mwe} % to get dummy images

\usepackage{hyperref}
\hypersetup{hypertex=true,
            colorlinks=true,
            linkcolor=blue,
            anchorcolor=blue,
            citecolor=blue}

\title{Advancing Cross-Organ Domain Generalization with Test-Time Style Transfer and Diversity Enhancement}
\name{Biwen Meng$^{1}$\textsuperscript{*}, Xi Long$^{1}$\textsuperscript{*}, Wanrong Yang$^{2}$, Ruochen Liu$^{1}$, Yi Tian$^{1}$, Yalin Zheng$^{3}$, Jingxin Liu$^{1}$\textsuperscript{\textdagger}\thanks{$^{2}$ Department of Computer Sciences; $^{3}$ Department of Eye and Vision Sciences; \textsuperscript{*} Equal Contribution; \textsuperscript{\textdagger} Corresponding author.}}

\address{$^{1}$ School of AI and Advanced Computing, Xi'an Jiaotong-Liverpool University, Suzhou, China; \\ 
$^{2, 3}$ University of Liverpool, Liverpool, United Kingdom}
\begin{document}
%\ninept
%
\maketitle
\begin{abstract}
Deep learning has made significant progress in addressing challenges in various fields including computational pathology (CPath). However, due to the complexity of the domain shift problem, the performance of existing models will degrade, especially when it comes to multi-domain or cross-domain tasks. In this paper, we propose a Test-time style transfer (T3s) that uses a bidirectional mapping mechanism to project the features of the source and target domains into a unified feature space, enhancing the generalization ability of the model. To further increase the style expression space, we introduce a Cross-domain style diversification module (CSDM) to ensure the orthogonality between style bases. In addition, data augmentation and low-rank adaptation techniques are used to improve feature alignment and sensitivity, enabling the model to adapt to multi-domain inputs effectively. Our method has demonstrated effectiveness on three unseen datasets. 
\end{abstract}
\section{Introduction}
\label{sec:intro}
%Machine learning (ML) has significantly contributed to natural language processing, computer vision, and healthcare. In computational pathology (CPath), ML is used to extract meaningful information from diverse types of pathology data, especially whole slide images (WSI) of tissue samples\cite{jahanifar2023domain}. Typically, machine learning algorithms learn from a training data set to build a model that performs tasks on target data\cite{cho2023complementary}. There is an assumption that the source data and target data are identically and independently distributed\cite{hastie2009elements}. However, this assumption is an idealized situation that is not valid in real scenarios. Data often comes from different distributions in the real world, leading to domain shift problems\cite{stacke2020measuring}. The performance will drop significantly when the model handles target domains outside of distribution. Therefore, it is necessary to enhance the generalization capabilities of machine learning models for unseen data.

Deep learning has made substantial advancements in various domains, including computer vision, natural language processing, and healthcare. In computational pathology (CPath), these models analyze whole slide images (WSIs) of tissue samples to extract critical insights from complex data \cite{jahanifar2023domain}. Typically, deep learning models are trained on labeled datasets under the assumption that training and test data share the same distribution \cite{cho2023complementary}. However, this assumption often fails in real-world scenarios, where domain shifts can degrade model performance when applied to new or unseen domains \cite{stacke2020measuring}. Addressing this challenge is essential for enhancing the generalization capabilities of deep learning models in diverse medical imaging tasks.

%Domain generalization (DG) is a method to enhance the ability of a model to generalize to unseen domains. It trains the model on one or several different but related domains to perform well on unseen test domains\cite{wang2022generalizing}. However, existing domain generalization methods (e.g., some adversarial learning, meta-learning) usually focus on learning domain-invariant features during training, which may not guarantee generalization to unseen data that differs greatly from the source distribution\cite{zhou2024test}. Recent research has explored test-time domain generalization (TTDG) frameworks, which leverage test data to improve model generalization without further training\cite{zhou2024test}. A promising approach is Test-Time Style Projection (TTSP), which has been successfully applied to tasks like Face Anti-Spoofing (FAS) by dynamically adjusting the style of test data to align it with known domains. Recent benchmarks indicate that many methods show minimal gains over basic supervised learning (ERM) in real-world conditions, implying that the current approaches may be overly complex and a new strategy is needed\cite{iwasawa2021test}.

\begin{figure}[ht]
    \centering
    \includegraphics[width=0.9\linewidth]{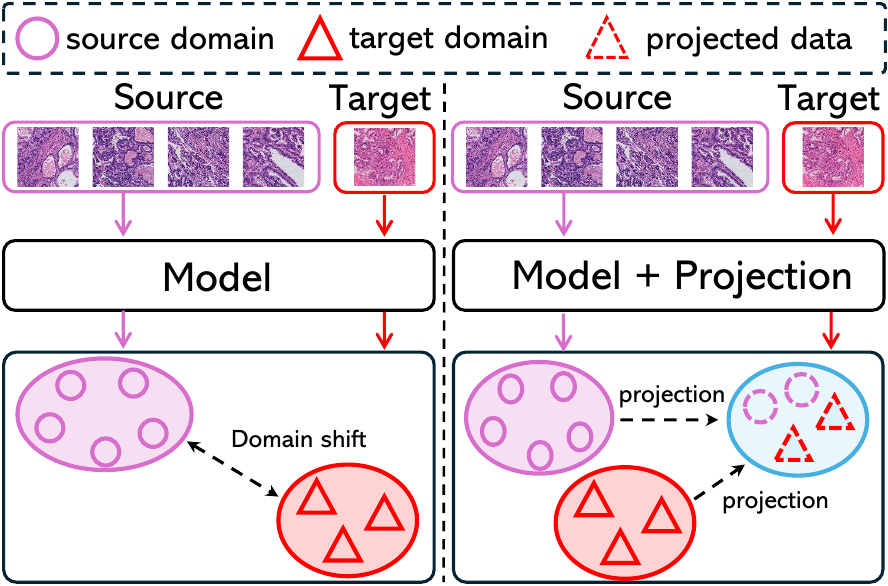}
     \caption{\textbf{Left}: Previous DG method, which learns domain-invariant features during training; \textbf{Right}: Test-time Style Transfer method, which projects the source domain and unseen domain into a style representation space, thereby enhancing model generalization on Cross-organ tasks.}
    \label{Figure 0}
\end{figure}

Domain generalization (DG) aims to improve a model's ability to perform well on unseen domains by training it on multiple related domains \cite{wang2022generalizing}. However, many existing DG methods, including adversarial learning and meta-learning approaches, primarily focus on learning domain-invariant features during training, leading to a huge computational burden\cite{sotto2024survey}. This focus may not ensure robust generalization to unseen data that significantly diverges from the source distribution \cite{zhou2024test}. Recent advancements have introduced test-time domain generalization (TTDG) frameworks, which utilize test data to enhance model generalization without additional training \cite{zhou2024test}. One effective strategy is Test-Time Style Projection (TTSP), which dynamically adjusts the style of test data to match known domains. It has been successfully applied in tasks like Face Anti-Spoofing (FAS). Nevertheless, recent benchmarks reveal that many of these methods yield only marginal improvements over basic supervised learning in real-world scenarios, indicating a need for simpler and more effective strategies \cite{iwasawa2021test}. Recently, foundation models (FMs) \cite{benigmim2024collaborating}, particularly in the histopathology domain, have shown strong potential for domain generalization due to their ability to leverage large-scale data for more effective adaptation across diverse domains.

\begin{figure*}[th]
    \centering
    \includegraphics[width=\textwidth, trim={0cm 3cm 0cm 2.5cm}, clip]{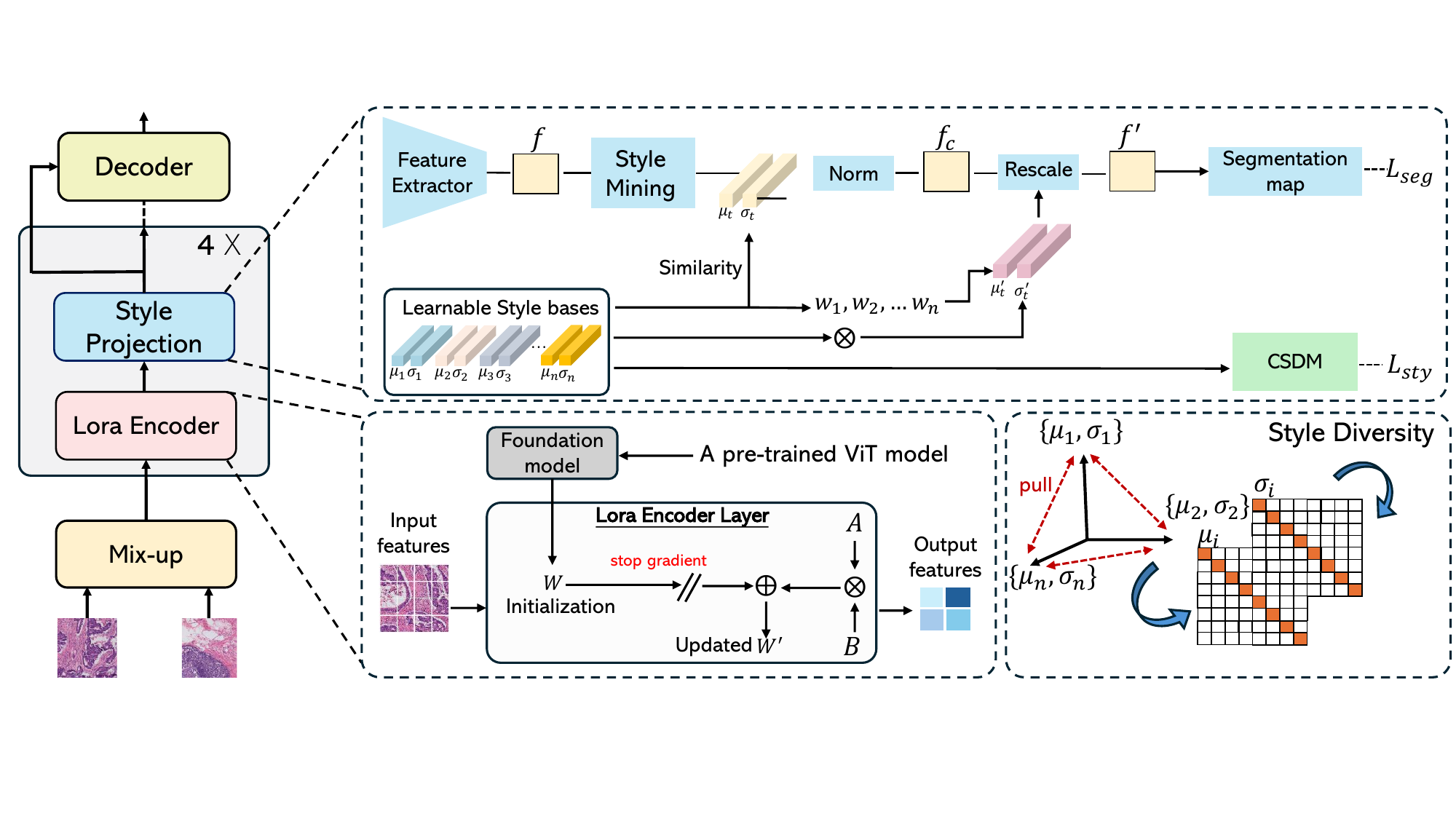}
    \caption{An overview of Test-Time Style Transfer (T3s) framework for Cross-organ tasks. First, we leverage pre-trained foundation models to extract robust, domain-independent features across domains. During training, we introduce Test-Time Style Transfer, a bidirectional mapping mechanism that maps the source and target domains into a style representation space. We then design a Cross-domain Style diversification module (CSDM) to maximize the style representation space. At test time, a domain adaptation strategy dynamically adjusts the style of the test data to align it with the learned features, thereby achieving effective generalization to unknown domains.}
    \label{fig-1}
\end{figure*}

In this paper, we propose Test-time Style Transfer (T3s), a foundation model-based approach that transfers test data into a style base to adapt to cross-organ domain generalization tasks \cite{chen2022scaling, juyal2024pluto}. Unlike traditional unsupervised domain generalization and test-time methods \cite{cho2023complementary, iwasawa2021test, zhou2024test}, T3s leverages the strong generalization capabilities of foundation models to capture domain-invariant features across different organs, as shown in Fig.\ref{Figure 0}. T3s introduces a bidirectional mapping mechanism, mapping both source and target domains into a style representation space, ensuring optimal feature mapping via learnable style bases. Instead of randomly selecting style bases, T3s combines multiple bases during training to find the most suitable combination. To further enhance performance, we propose a Cross-domain Style Diversification Module (CSDM) to ensure orthogonality between style bases, maximizing the representation space. Data augmentation and low-rank adaptation techniques are also employed to improve feature alignment, allowing the model to effectively handle inputs from multiple domains. Experiments demonstrate the effectiveness and superiority of T3s. The contributions of this paper are two-fold:

\begin{itemize}
   \item We propose Test-time Style Transfer (T3s), a novel method that bi-directionally approximates style features by projecting the test domain and the source domain into the style representation space.
\end{itemize}
\begin{itemize}
    \item We integrate the Cross-domain Style Diversification Module (CSDM) into the foundation model, which is the first domain generalization model that focuses on the morphological differences caused by different organs.
\end{itemize}
\begin{itemize}
    \item Our proposed method outperforms state-of-the-art methods in cross-organ generalization tasks based on experimental results.
\end{itemize}

%why here? delete?\textbf{Key Difference}: T3s ensures consistency between the training and inference processes. Specifically, during the training phase, we do not randomly select a target style, but directly query the style base and combine multiple bases to construct a target style. The model learns to map the input features to this combined style. This means that the model not only learns style transfer during training but also learns how to effectively query and combine multiple style bases, ensuring that it can find the most suitable style combination in practical applications. 

%\textbf{Related work:}  Test-time adaptation or TTDG is closely related to our work. Test-time adaptation proposes full test-time adaptation to adjust parameters by minimizing the prediction entropy using stochastic gradient descent\cite{iwasawa2021test}. TTDG transfers the style of the test sample to the nearest source domain before making predictions\cite{wang2020tent}. However, test-time methods have not been widely studied in the field of domain generalization, and content loss is challenging on WSI. In particular, in domain generalization tasks, content loss is often insufficient to handle the variability between the source and target domains\cite{sanakoyeu2018style,peng2024cyclic}.

%These guidelines include complete descriptions of the fonts, spacing, and
%related information for producing your proceedings manuscripts.

\section{Method}
\label{sec:format}

%All printed material, including text, illustrations, and charts, must be kept
%within a print area of 7 inches (178 mm) wide by 9 inches (229 mm) high. Do
%not write or print anything outside the print area. The top margin must be 1
%inch (25 mm), except for the title page, and the left margin must be 0.75 inch
%(19 mm).  All {\it text} must be in a two-column format. Columns are to be 3.39
%inches (86 mm) wide, with a 0.24 inch (6 mm) space between them. Text must be
%fully justified.h

This paper proposes a new method to address the domain shift problem between the training dataset and the test dataset. Given a target domain $\mathcal{T}$, the domain shift $\epsilon_{\mathcal{T}}(h)$ is typically defined as the prediction error between the ideal segmentation function $h^*_{\mathcal{T}}(x)$ and the segmentation model $h(x)$, which is denoted as $\epsilon_{\mathcal{T}}(h) = \left| h(x) - h^*_{\mathcal{T}}(x) \right|$ \cite{ben2010theory}. Given that the training dataset $\Lambda_\mathcal{S} = {\{ \mathcal{S}_i \mid i = 1, 2, \dots, n \}}$ consists of $n$ different source domains, the domain shift $\epsilon_{\mathcal{T}}(h)$ is upper-bounded \cite{albuquerque2019generalizing} by:
\begin{equation}
    \epsilon_{\mathcal{T}}(h) \leq \frac{\rho}{2}+ \frac{\gamma}{2} + \sum^n_{i}{\eta^*_i\epsilon_{\mathcal{S}_i}(h)} \label{eq:risk}
    % \lambda_{\mathcal{H}, \left(\mathcal{D}_\mathcal{T}, \mathcal{D}^*_\mathcal{S}\right)} 
\end{equation}
where $\rho$ denotes as largest domain shifts between each pair of source domains distribution, which can be measured by $\mathcal{H}$-divergence\cite{ben2010theory}: 
\begin{equation}
    \rho = \sup_{\mathcal{S}_i,\mathcal{S}_j \in \Lambda_\mathcal{S}}d_\mathcal{H}(\mathcal{D}_{\mathcal{S}_i}, \mathcal{D}_{\mathcal{S}_j})
\end{equation}
$\gamma$ denotes domain shifts between target domain distribution and source domain distribution, which can be also measured by $\mathcal{H}$-divergence\cite{ben2010theory}:
\begin{equation}
    \gamma \iffalse  =  \min d_\mathcal{H}\left(\mathcal{D}_\mathcal{T}, \Lambda_\mathcal{S}\right) \fi = \min  d_\mathcal{H}\left(\mathcal{D}_\mathcal{T}, \sum_{i}^{n}\eta_i\mathcal{D}_{\mathcal{S}_i}\right)
\end{equation}
 The weights $\eta^*$ are defined as $\mathop{\arg\min} \limits_{\eta \iffalse \in \triangle_{n} \fi} d_\mathcal{H}\left(\mathcal{D}_\mathcal{T}, \sum_{i}^{n}\eta_i\mathcal{D}_{\mathcal{S}_i}\right)$, denoting the minimization of the domain shift $\gamma$.%, and $\mathcal{D}^*_\mathcal{S}$ is the best approximate within a convex hull $\Lambda_\mathcal{S} = \left\{ \sum_{i}^{n}\eta_i\mathcal{D}_{\mathcal{S}_i}\mid \eta \in \mathbb{R}^n \right\}$ of training datasets.
%Finally, $\lambda_{\mathcal{H}, \left(\mathcal{D}_\mathcal{T}, \mathcal{D}^*_\mathcal{S}\right)}$ denotes as the ideal joint risk across the target domain and the domain with the best approximate distribution $\mathcal{D}^*_\mathcal{S}$ \cite{Albuquerque2019AdversarialTR}.

% Fig.\ref{fig-1} presents an overview of the proposed Test-Time Style Transfer (T3s) framework. Our method integrates three core strategies to minimize domain shift: a foundation model, data augmentation, and test-time domain adaptation. First, we leverage a pre-trained foundation model to extract robust, domain-independent features across different domains. During training, advanced data augmentation techniques mix features from various organizational domains, enhancing data diversity and promoting the learning of more generalized representations. Finally, at test time, a domain adaptation strategy dynamically adjusts the test data's style to align with the learned features, enabling effective generalization to unseen domains.

As shown in Fig.\ref{fig-1}, our method integrates three core strategies to minimize domain shift $\epsilon_{\mathcal{T}}(h)$: a foundation model, data augmentation, and test-time domain adaptation.

\subsection{Foundational Model}
% [4] Collaborating Foundation Models for Domain Generalized Semantic Segmentation

% [5] Do Vision Foundation Models Enhance Domain Generalization in Medical Image Segmentation

% [6] Scaling Vision Transformers to Gigapixel Images via Hierarchical Self-Supervised Learning

Foundation models pre-trained on large datasets using self-supervised learning have demonstrated the ability to extract generalized features, significantly improving downstream task performance \cite{benigmim2024collaborating}, formulated as the third term $\sum_{i=1}^{n} \eta^*_i\epsilon_{\mathcal{S}_i}(h)$ in Eq.\ref{eq:risk}. Leveraging this, we adopted a foundation model as the backbone for our segmentation approach, pre-trained on 104 million pathology image patches from TCGA using the DinoV2 framework.

\subsection{Domain Argumentation}
We used mixup for domain augmentation, a technique designed to enhance model robustness by generating synthetic training datasets potentially similar to the target domain $\mathcal{T}$, thereby reducing the domain shift  $\gamma$ in Eq.\ref{eq:risk} between the source and target domains. This can be formulated as follows:

\begin{equation}
\label{eq:mixup}
x_{new} = \lambda x_p + (1 - \lambda) x_q
\end{equation}
In this equation, $\lambda$ is a random variable sampled from a uniform distribution over the interval (0, 1), and $x_p$ and $x_q$ denote randomly selected images from the source domain dataset $\mathcal{S}$.

%As illustrated in Eq. $\ref{eq:mixup}$, mixup is an effective method compared to other augmentation techniques, such as color adjustment or noise injection. It not only generates new samples that may help reduce domain shifts $\gamma$ as shown in Eq. \ref{eq:risk}, but also ensures that the largest domain shifts $\rho$ remain unchanged, as indicated in Eq. \ref{eq:risk}.

\subsection{CSDM}
%This decomposition can be formulated as:
%\begin{equation}
%\left[\begin{matrix}
%f_c \\ fs
%\end{matrix}\right] = \Phi(f)
% %   f_c,f_s = \Phi(f)
%\end{equation}
According to Piratla et al. (2020) \cite{piratla2020efficientdomaingeneralizationcommonspecific}, image features $f$ can be decomposed into a specific feature  $f_s$ and a common feature  $f_c $, where  $f_c$  is shared across domains and directly related to the segmentation mask. Inspired by \cite{Karras2018ASG} and \cite{8237429}, we define the specific feature  $f_s$  as the style component, the common feature  $f_c$ corresponding to the content, the decomposition function $\Phi(f)$ can be expressed as:
\begin{equation}
%    f_s = {\mu}_s, {\sigma}_s 
\left[\begin{matrix}
f_s \\ f_c
\end{matrix}\right] = \Phi(f) =
    \left\{
    \begin{array}{cc}
        {\mu}_s, {\sigma}_s \\
        (f - \mu_s)/{\sigma_s}
    \end{array}
    \right.
\end{equation}
where:
\begin{equation}
    \mu_s = \frac{1}{HW}\sum_{h=1}^{H}\sum_{w=1}^{W}f, \sigma_s=\sqrt{\frac{1}{HW}\sum_{h=1}^{H}\sum_{w=1}^{W}(f-{\mu}_s)^2}
\end{equation}
Given the new style feature  $f^{\prime}_s$  and the content feature  $f_c$ , the segmentation loss  $L_{\text{seg}}$  is defined as the cross-entropy loss between the predicted segmentation mask and the ground truth mask. Formally, this can be written as:
\begin{equation}
    L_{\text{seg}} = CE(D(f^\prime), mask_{gt})
\end{equation}
Here,  $f^{\prime} = \Phi^{-1}(f^{\prime}s, f_c) =  \sigma_s^\prime f_c + \mu_s^\prime$ , where $D$  represents the segmentation decoder,  $mask_{gt}$  is the ground truth segmentation mask, and  $CE$  denotes the cross-entropy loss.

As shown in Fig.\ref{Figure 0}, our objective is to project the image features $f$ from both the target domains in the test datasets and the source domains in the training dataset into the same feature space as a weighted combination of bases. This approach not only has the potential to reduce the largest domain shift $\rho$ between each pair of source domains in Eq.\ref{eq:risk} but also decreases the domain shift $\gamma$ between the source and target domains. Since  $f_c$  is the common feature shared across all domains and is related to the segmentation mask, we only project the style feature  $f_s$  to ensure that the new feature  $f^{\prime} = \Phi^{-1}(f^{\prime}_s, f_c)$  does not degrade segmentation performance.
We define the learnable style bases as $B = \{B_i = (\mu_i, \sigma_i) \mid i = 1, 2, 3, \dots, n\}$, and the projection process is formulated as:
\begin{equation}
    f^{\prime}_s = \sum_{i=0}^{n} \frac{\exp{d_i}}{\sum_{j=0}^{n} \exp{d_j}} * B_i
\end{equation}
where the cosine distance vector $d_i \in \mathbb{R}^2$  is defined as:
\begin{equation}
    d_i = \frac{f_s \cdot B_i}{|f_s||B_i|} = \left({\frac{\mu_s \cdot \mu_i}{|\mu_s||\mu_i|}}, {\frac{\sigma_s \cdot \sigma_i}{|\sigma_s||\sigma_i|}}\right)
\end{equation}

To ensure that all features can be projected into the style representation space as a weighted combination of style bases and that the style bases fully represent the style space, we aim to make the style bases as orthogonal as possible. The loss function is defined as:

\begin{equation}
L_{\text{sty}} = \frac{1}{(n-1)^2}\sum_{i=1}^{n} (1-\delta_{ij}) \left(\frac{B_i \cdot B_j}{|B_i||B_j|}\right)^2
\end{equation}
where  $\delta_{ij}$  is the Kronecker delta.

%[10] EFFICIENT DOMAIN GENERALIZATION VIA COMMON-SPECIFIC LOW-RANK DECOMPOSITION

\section{Experiment}
\label{sec:pagestyle}
\subsection{Dataset}
In this experiment, we used the Patch dataset (180 source domains and 120 target domains) collected from Shanghai Ruijin Hospital from 186 WSI adenocarcinoma datasets of different tissues, which were marked as training sets in the COSAS2024 Challenge \footnote{https://cosas.grand-challenge.org/}. The details are shown in Table \ref{tab1:dataset}.
\begin{table}[h]
\centering
\caption{Dataset of Experiment}
\begin{tabular}{c|p{2cm}|p{2cm}}
\toprule
\diagbox[width=2.5cm,height=1.0cm]{Domains}{Organs}& \makecell[c]{Stomach \\ Pancreas \\ Colorectum} & \makecell[c]{Ampullary \\ Gallbladder \\ Intestine} \\
%       & Stomach Pancreas Colorectum  & Ampullary Gallbladder Intestine\\
\midrule
Source Domain &  60 & -\\
\midrule
Target Domain & 20 & 20 \\
\bottomrule
\end{tabular}
\label{tab1:dataset}
\end{table}

\subsection{Implementation Details}
We use ViT-Small as the backbone for feature extraction and UPerNet as the decoder for segmentation. Training is performed with a batch size of 8, patch size of 512, and AdamW optimizer over 320,000 iterations. Model performance is evaluated using intersection over union (IoU) and Dice coefficient, with higher values indicating better results.
%We employ ViT-Small as the backbone to extract features from the input image, paired with UPerNet as the decoder to generate segmentation maps. The training is conducted with a batch size of 8 and a patch size of 512, using the AdamW optimizer for 320,000 iterations. To evaluate model performance, we use the intersection over union (IoU) and Dice coefficient metrics - higher values for both indicate better performance.

\subsection{Result}
As shown in Table \ref{tab2:map-domain}, we compared our proposed method with state-of-the-art methods, including Vit\cite{thisanke2023semanticsegmentationusingvision},  Swin\cite{10.1145/3691521.3691537}, CFM\cite{benigmim2024collaboratingfoundationmodelsdomain}, DCAC \cite{article} and TTDG \cite{zhou2024test}. From the table, Our method achieves the highest IoU (77.68\%) and Dice coefficient (87.43\%), outperforming other state-of-the-art models like TTDG (76.90\%, 86.94\%) and DCAC (76.61\%, 86.75\%). ViT and CFM perform slightly lower, with Dice scores of 78.49\% and 85.44\%. These improvements are due to our innovative architecture and advanced training strategies, which enhance accuracy and generalization, especially for cross-organs segmentation tasks.
\begin{table}[h]
\centering
\caption{Comparison with state-of-the-art methods.}
\begin{tabular}{c|cc}
\toprule
    Method           &   IOU (\%)  $\uparrow$ & DICE (\%) \\
\midrule
ViT\cite{thisanke2023semanticsegmentationusingvision} & 64.58 & 78.49 \\
Swin\cite{10.1145/3691521.3691537}  & 76.77 & 86.86 \\
CFM\cite{benigmim2024collaboratingfoundationmodelsdomain} & 74.58 & 85.44 \\
DCAC \cite{article} & 76.61 & 86.75 \\ 
TTDG \cite{zhou2024test} & 76.90 &  86.94\\
\textbf{Ours}    & \textbf{77.68} & \textbf{87.43} \\  
\bottomrule
\end{tabular}
\label{tab2:map-domain}
\end{table}
%[1]
%[2] Domain and content adaptive convolution based multi-source domain generalization for medical image segmentation
%[3] Collaborating Foundation Models for Domain GeneralizedSemantic Segmentation

\subsection{Ablation Study}
We conducted an ablation study to validate the necessity of the three modules in T3s: Fine-Tune, Mixup, and CSDM. As shown in Table \ref{tab:ablation}, all combinations of fine-tuning, data augmentation, and CSDM were compared. Firstly, the model performance was significantly jump after fine-tuning and data augmentation. Furthermore, after adding CSDM, the model performance was further improved, proving the effectiveness of our CSDM in the multi-domain generalization task.

\begin{table}[h]
\centering
\caption{Ablation on the effectiveness of various components, including Mixup, Foundational Model (FM), and Cross-domain Style Module (CSDM).}
\label{tab:ablation}
{\footnotesize
\resizebox{0.49\textwidth}{!}{%
\begin{tabular}{ccc|cc}
\toprule
FM & Mixup & CSDM &  IOU (\%)  $\uparrow$ & DICE (\%)  $\uparrow$\\
\midrule
& & & 64.58 & 78.49 \\
\checkmark & & & 76.32 & 86.57  \\
 & \checkmark & & 64.75 & 78.60 \\
  & & \checkmark & 64.09 & 78.12 \\
& \checkmark & \checkmark & 63.47 & 77.65  \\
\checkmark & \checkmark &  & 77.41 & 87.25 \\
\checkmark & & \checkmark & 76.65 & 86.78 \\
\checkmark & \checkmark & \checkmark & \textbf{77.68} & \textbf{87.43} \\
\bottomrule
\end{tabular}%
}
}
\end{table}

\subsection{Visualization}
\begin{figure}[htbp]
\centering 
\subfigure[Before Style Projection]{
\includegraphics[height=3.0cm]{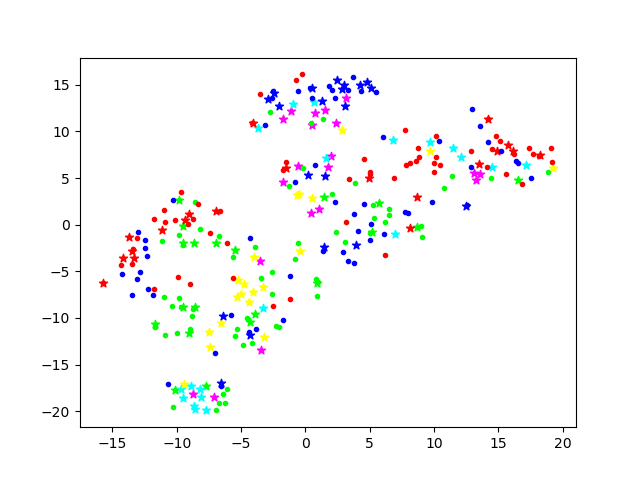}
}
\subfigure[After Style Projection]{
\includegraphics[height=3.0cm]{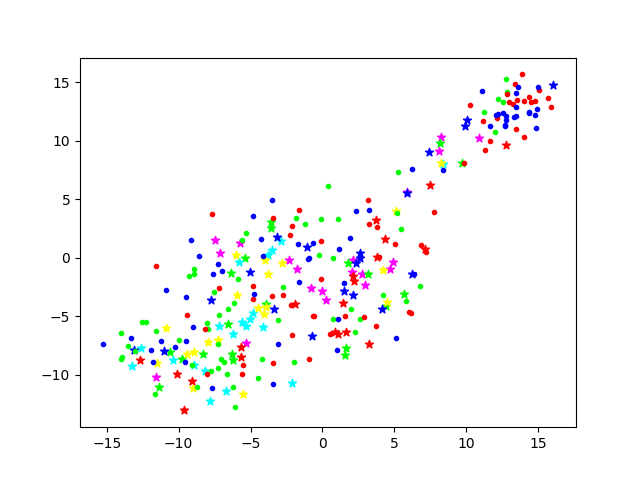}
}
\caption{Features of different domains are visualized with T-SNE before (a) and after (b) test-time style projection. Training domain: \textcolor[rgb]{1.0, 0, 0}{$\bullet$} Pancreas, \textcolor[rgb]{0, 1.0, 0}{$\bullet$}  Colorectum, \textcolor[rgb]{0, 0.0, 1.0}{$\bullet$} Stomach; Testing domain: \textcolor[rgb]{1.0, 0, 0}{$\star$} Pancreas, \textcolor[rgb]{0, 1.0, 0}{$\star$}  Colorectum, \textcolor[rgb]{0, 0.0, 1.0}{$\star$} Stomach, \textcolor[rgb]{1.0, 1.0, 0}{$\star$}  Intestine, \textcolor[rgb]{1.0, 0, 1.0}{$\star$}  Ampullary, \textcolor[rgb]{0, 1.0, 1.0}{$\star$}  Gallbladder}
\label{fig:tsne}
\end{figure}

Fig.\ref{fig:tsne} illustrates style distribution changes across domains before and after style projection. In Fig.\ref{fig:tsne} (a), before style projection, the feature distribution of different organ domains is relatively scattered. After projection (Fig.\ref{fig:tsne} (b), the style distribution of the unseen domain is aligned to the learnable style space, which is closer to the source domain. This shows that T3s can effectively project the style of unseen domains, reduce inter-domain differences, and improve the generalization ability of the model.

%\subfigure[DCAC]{
%		\centering
%		\includegraphics[height=5.0cm]{tsne/swin-unet-b0-8xb2-160k-cosas-512x512.png}
%	}
%	\subfigure[TTDG]{
%		\centering
%		\includegraphics[height=5.0cm]{tsne/2-vit-lora-pretained-ttsp11-512x512-task1-b8-320k.png}
%	}
%	\subfigure[Ours]{
%		\centering
%		\includegraphics[height=5.0cm]{tsne/0-vit-lora-pretained-mixup-ttsp11-512x512-task1-b8-320k.png}
%        }
%	\caption{Feature visualization by t-SNE. Training domain: \textcolor[rgb]{1.0, 0, 0}{$\bullet$} Pancreas, \textcolor[rgb]{0, 1.0, 0}{$\bullet$}  Colorectum, \textcolor[rgb]{0, 0.0, 1.0}{$\bullet$} Stomach; Testing domain: \textcolor[rgb]{1.0, 0, 0}{$\star$} Pancreas, \textcolor[rgb]{0, 1.0, 0}{$\star$}  Colorectum, \textcolor[rgb]{0, 0.0, 1.0}{$\star$} Stomach, \textcolor[rgb]{1.0, 1.0, 0}{$\star$}  Intestine, \textcolor[rgb]{0, 1.0, 0}{$\star$}  Ampullary, \textcolor[rgb]{0, 1.0, 0}{$\star$}  Gallbladder}
%	\label{fig:tsne}
%\end{figure*}
\section{Conclusion}
\label{sec:typestyle}
We propose T3s, a novel framework for cross-organ domain generalization, featuring bidirectional style projection between target and source domains. T3s leverages a Foundation Model with test-time style transfer and a Cross-domain Style Diversity Module (CSDM) to enhance generalization by aligning styles and expanding the style space. Future work will explore extending T3s to other medical imaging tasks and advancing domain generalization techniques.
%In this paper, we propose an innovative approach to improve the domain generalization ability of the model through the bidirectional projection of the target domain and source domain, marking the first work on cross-organ domain generalization. Our framework T3s integrates with the Foundation Model, using test-time style transfer to align unseen sample styles and a cross-domain style diversity module (CSDM) to expand style space with cosine orthogonality, further improving generalization. Future work will extend T3s to broader tasks to maximize their potential.

\section{Acknowledgement}
This work was jointly supported by the National Natural Science Foundation of China (62201474), Suzhou Science and Technology Development Planning Programme (Grant No.ZXL2023171) and XJTLU Research Development Fund (RDF-21-02-084).

%To achieve the best rendering both in the printed and digital proceedings, we strongly encourage you to use Times-Roman font.  In addition, this will give the proceedings a more uniform look.  Use a font that is no smaller than nine point type throughout the paper, including figure captions. In nine point type font, capital letters are 2 mm high.  If you use the smallest point size, there should be no more than 3.2 lines/cm (8 lines/inch) vertically.  This is a minimum spacing; 2.75 lines/cm (7 lines/inch) will make the paper much more readable.  Larger type sizes require correspondingly larger vertical spacing.  Please do not double-space your paper.  True-Type 1 fonts are preferred. The first paragraph in each section should not be indented, but all the following paragraphs within the section should be indented as these paragraphs demonstrate.

\bibliographystyle{IEEEbib}
\bibliography{refs}

\end{document}